\documentclass[letterpaper, 10 pt, conference]{ieeeconf}
\IEEEoverridecommandlockouts
\usepackage{cite}
\usepackage{amsmath,amssymb,amsfonts}
\usepackage{algorithm,algpseudocode}
\usepackage{soul}
\usepackage{graphicx}
\usepackage{subcaption}
\usepackage{textcomp}
\usepackage{xcolor}
\usepackage{url}
\usepackage{booktabs} 
\def\BibTeX{{\rm B\kern-.05em{\sc i\kern-.025em b}\kern-.08em
    T\kern-.1667em\lower.7ex\hbox{E}\kern-.125emX}}
\begin{document}

\title{\LARGE \bf
Directed-MAML: Meta Reinforcement Learning Algorithm with Task-directed Approximation
}


\author{Yang Zhang, Huiwen Yan and Mushuang Liu 
\thanks{This work was supported by DARPA Young Faculty Award with the grant number D24AP00321.}
\thanks{Yang Zhang is with the Department of Mechanical and and Aerospace Engineering , University of Missouri, Columbia, MO, USA
{\tt\small zhangy1@missouri.edu}}%
\thanks{Huiwen Yan and Mushuang Liu are with the Department of Mechanical Engineering, Virginia Polytechnic Institute and State University, Blacksburg, VA, USA
{\tt\small huiweny@vt.edu, mushuang@vt.edu}}%
}

\maketitle

\begin{abstract}
Model-Agnostic Meta-Learning (MAML) is a versatile meta-learning framework applicable to both supervised learning and reinforcement learning (RL). However, applying MAML to meta-reinforcement learning (meta-RL) presents notable challenges. First, MAML relies on second-order gradient computations, leading to significant computational and memory overhead. Second, the nested structure of optimization increases the problem's complexity, making convergence to a global optimum more challenging. To overcome these limitations, we propose Directed-MAML, a novel task-directed meta-RL algorithm. Before the second-order gradient step, Directed-MAML applies an additional first-order task-directed approximation to estimate the effect of second-order gradients, thereby accelerating convergence to the optimum and reducing computational cost. Experimental results demonstrate that Directed-MAML surpasses MAML-based baselines in computational efficiency and convergence speed in the scenarios of CartPole-v1,  LunarLander-v2 and two-vehicle intersection crossing. Furthermore, we show that task-directed approximation can be effectively integrated into other meta-learning algorithms, such as First-Order Model-Agnostic Meta-Learning (FOMAML) and Meta Stochastic Gradient Descent(Meta-SGD), yielding improved computational efficiency and convergence speed.
\end{abstract}


\section{Introduction}
Training deep neural networks typically requires a large volume of data to effectively capture the underlying characteristics of the data distribution. A well-trained model is expected to generalize to unseen data and perform comparably to its performance in the training set. However, when the amount of training data is limited, the model may struggle to learn a representative data distribution, resulting in poor generalization and degraded performance on unseen samples \cite{zhang2017understanding,shorten2019survey}.  As a solution, meta-learning, also known as learning to learn, is formulated as a framework that enables the model to adapt quickly to new tasks with a small amount of data. Thus, it can be widely applied to various tasks such as computer vision and natural language processing \cite{DBLP:journals/corr/abs-1909-13579,DBLP:journals/corr/abs-2007-09604,DBLP:journals/corr/abs-2009-08445,Gui_2018_ECCV,Elsken_2020_CVPR,lee-etal-2021-meta,Wang_2019_ICCV,Wang_2022_CVPR}.

Among various meta-learning algorithms, MAML \cite{MAML} stands out as a widely used optimization-based framework, particularly effective in few-shot learning scenarios. The core idea of MAML is to learn a set of meta-parameters that may not be optimal for any single task but are well-initialized for rapid adaptation across many tasks. These meta-parameters are optimized to enable fast convergence to task-specific solutions with a few gradient steps. MAML has demonstrated strong performance in data-scarce domains, including computer vision and language modeling \cite{mamlcv1,mamlcv2,mamlnlp1}.

 However, MAML also faces several limitations that hinder its effectiveness in meta-RL tasks. First, the outer-loop update in MAML involves computing second-order gradients, which is computationally intensive and time-consuming especially when optimizing across multiple tasks simultaneously. Although first-order approximations (e.g., FOMAML) can alleviate this issue, they often lead to slower convergence and still require aggregating gradients from multiple tasks. Second, MAML poses a nested structure of optimization problem \cite{sharpMAML}, making it prone to challenges such as saddle points and local optima, which can hinder convergence to the global optimum. The sparse and delayed rewards in meta-RL aggregate this problem. As a result, hyperparameters such as the inner and outer step sizes and the number of inner gradient steps must be carefully tuned to ensure the optimization towards global optimum. 

To address above challenges, we propose Directed-MAML, a meta-RL algorithm that incorporates a task-directed approximation strategy. The core idea is to introduce a cross-task pre-adaptation step into the standard MAML framework. Specifically, before performing the typical inner-loop adaptations and outer-loop meta-updates, the algorithm identifies a medium task by averaging environment parameters across the task distribution. This task is hypothesized to influence the meta-gradient direction. Directed-MAML then performs a first-order gradient update using trajectories sampled from this representative task to approximate the effect of second-order gradients. Since computing the first-order gradient for a single task requires fewer resources than computing second-order gradients across multiple tasks, Directed-MAML improves the model’s computational efficiency, defined as the amount of computation required to reach the global optimum.

There are mainly two aspects of contributions of this paper.

\begin{itemize}
\item Directed-MAML, a task-directed meta-reinforcement learning algorithm, is introduced to enhance training efficiency. Experimental results demonstrate that Directed-MAML accelerates the convergence of meta-training and reduces the computational resources required to reach the global optimum.
\item We propose a model-agnostic task-directed approximation strategy that enhances the training efficiency of gradient-based meta-learning algorithms. This approach is broadly applicable and can be integrated into various MAML-style methods. We demonstrate that incorporating task-directed approximation into FOMAML and Meta-SGD leads to improved computational efficiency for training.

\end{itemize}

\section{Related works}

We review the related works in three aspects.

\textbf{Reinforcement learning.} RL solves sequential decision-making problems \cite{sutton2018reinforcement}. In RL, an agent learns to maximize the cumulative reward through interactions with the environment. Among value-based methods, Q-learning \cite{qlearning} is a classic algorithm that learns the optimal action-value function through temporal-difference updates. Deep Q-Networks (DQN) \cite{dqn} extend Q-learning by using deep neural networks as function approximators, achieving outstanding performance on high-dimensional tasks such as Atari games \cite{dqn}. REINFORCE algorithm, a policy-based method, directly optimizes the policy by estimating the gradient of the expected return. Actor-critic algorithms \cite{ac} combine the advantages of both value-based and policy-based approaches by learning both a policy (actor) and a value function (critic), offering improved sample efficiency and stability. These algorithms have laid the foundation for RL applications and continue to serve as the basis for more advanced and domain-specific approaches.

\textbf{Meta-reinforcement learning.} In meta-RL, the goal is to enable agents to quickly adapt to new tasks by leveraging experiences across a task distribution. Reinforcement Learning Squared (RL²) \cite{rl2} achieves this by training recurrent policies to adapt online through hidden state updates, effectively learning a learning algorithm. Simple Neural Attentive Meta-Learner (SNAIL) \cite{snail} incorporates temporal convolutions and attention mechanisms into a meta-learner to improve adaptation efficiency. Probabilistic Embeddings for Actor-Critic Reinforcement Learning (PEARL) \cite{pearl} introduces probabilistic context variables for task inference, enabling a sample-efficient off-policy meta-RL. These approaches offer diverse perspectives on improving task adaptation and sample efficiency in meta-RL. Our work builds on this direction by incorporating task-directed signals to enhance computational efficiency during meta-policy training.

\textbf{Model-agnostic meta learning.} MAML \cite{MAML} is a widely adopted meta-learning algorithm that learns meta policy parameters capable of fast adaptation to new tasks with limited gradient steps and data. This algorithm can also be applied to meta-RL problems. While effective in both supervised and reinforcement learning settings, MAML’s reliance on second-order gradients leads to high computational cost and potential instability \cite{mamlplus}. To improve efficiency, first-order variants such as FOMAML \cite{FOMAML} and Reptile \cite{reptile} approximate or avoid second-order computations. Meta-SGD (Li et al., 2017) extends MAML by learning both initialization and learning rates, enabling faster task adaptation. Sharp-MAML \cite{sharpMAML} introduces the idea of learning a flat and robust loss landscape around the initialization point by integrating sharpness-aware optimization into MAML, which improves generalization and stability during adaptation. Additional improvements include MAML++ \cite{mamlplus} and iMAML \cite{iMAML}, where the former one stabilizes training with optimization refinements and the latter one employs implicit differentiation for improved bi-level optimization. Based on above works, our method proposes a task-directed approximation strategy that enhances computational efficiency while preserving the model-agnostic nature of MAML. This paper focuses on improving the computational efficiency and reducing the number of training epochs required for convergence in models trained with MAML by incorporating task-directed approximation.

\section{Preliminary and Motivation}

In this section, we provide a brief overview of MAML on meta-RL and then discuss the optimization challenges associated with training MAML-based models on RL problems.

\subsection{Problem Formulation of MAML on meta-RL}

MAML aims to train models that can rapidly adapt to new tasks using only a small number of samples, typically ranging from one to five samples per task. Consider a task distribution $\rho(\mathcal{T})$, where $\mathcal{T}_i$ denotes a task indexed by $i$. Each task is modeled with a Markov Decision Process (MDP) and each MDP is denoted as a tuple: $\mathcal{T}_i = (\mathcal{S}, \mathcal{A}, \mathcal{P}_i, \mathcal{R}_i,  \gamma)$, and $\mathcal{S}$ denotes the state space and $\mathcal{A}$ denotes the action space. The transition model is denoted as $\mathcal{P}_i: \mathcal{S} \times \mathcal{A} \times \mathcal{S} \rightarrow (0,1]$, while the reward function is denoted by $\mathcal{R}_i: \mathcal{S} \times \mathcal{A} \rightarrow \mathbb{R}$. The discount factor is given by $\gamma \in (0,1]$. Note that $\mathcal{S}$, $\mathcal{A}$ and $\gamma$ have no subscripts because they are shared across tasks. The policy of the agent is denoted as $\pi:\mathcal{S}\rightarrow\Delta(\mathcal{A})$, where $\Delta(\mathcal{A})$ is the probability simplex over action space $\mathcal{A}$. Meta policy, the higher-level policy learned over a distribution of tasks, is parameterized with $\theta$ while the policy of each task is parameterized with $\theta'_i$. 


MAML solves the following optimization problem:

\begin{subequations}

    \begin{equation}
    \max_{\theta}\sum_{\mathcal{T}_i \sim \rho(\mathcal{T})}\mathcal{J}_{\mathcal{T}_i}(\theta'_i) = \max_{\theta}\sum_{\mathcal{T}_i \sim \rho(\mathcal{T})}\mathcal{J}_{\mathcal{T}_i}(\theta + \alpha \nabla_\theta \mathcal{J}_{\mathcal{T}_i}(\theta)) , 
        \end{equation}

\begin{equation}
\mathcal{J}_{\mathcal{T}_i}(\theta'_i) = \mathbb{E}_{s\sim \rho(s_0)}[\sum_{t=0}^\infty\gamma^t\mathcal{R}_i(s_t,a_t)|\pi_{\theta'_i}, s_0=s],
\end{equation}
    
\end{subequations}

\noindent where $\mathcal{J}_{\mathcal{T}_i}$ denotes the cumulative reward of task $\mathcal{T}_i$, and $\rho(s_0)$ denotes the distribution of initial states. The optimization objective of MAML is to find an optimal meta policy parameter $\theta$ which can maximize the sum of cumulative reward
$\mathcal{J}_{\mathcal{T}_i}$ across the task distribution $\rho(\mathcal{T})$.

There are two nested loops in MAML training process, inner loop and outer loop. The inner loop performs a small number of gradient steps on a specific task $\mathcal{T}_i$ while the outer loop updates the shared meta-parameters $\theta$ across tasks. The gradient steps of inner and outer loop are:

\begin{subequations}
\begin{equation}
    \theta_i' = \theta + \alpha \nabla_\theta \mathcal{J}_{\mathcal{T}_i}(\theta),
    \label{2a}
\end{equation}

\begin{equation}
    \theta^{(t+1)} = \theta^{(t)} + \beta \sum_{\mathcal{T}_i \sim \rho(\mathcal{T})} \nabla_\theta \mathcal{J}_{\mathcal{T}_i}
    (\theta_i'^{(t)}), 
    \label{2b}
\end{equation}
\end{subequations}

\noindent where $\alpha$ and $\beta$ denote the step size of inner and outer loops.

Meanwhile, MAML has a simplified version known as FOMAML \cite{FOMAML}, which improves computational efficiency by replacing the second-order gradient with a first-order approximation during the meta-update phase. The inner loop and outer loop of FOMAML are the same as MAML's (Equation \ref{2a} and \ref{2b}).

\subsection{Motivation of Directed-MAML}

Classic MAML has demonstrated strong performance across a wide range of supervised classification and regression tasks, confirming its effectiveness as a meta-learning framework. However, it still faces significant optimization challenges for meta-RL. As previously discussed, these challenges stem from the high computational cost of second-order gradient evaluations and the difficulty of achieving convergence to the global optimum. Directed-MAML is motivated by addressing these challenges on meta-RL. The main purpose of Directed-MAML is to find an approximation of MAML's second-order derivative which is more computationally efficient and easier to converge to optimum for meta-RL.

The meta-RL problem addressed by Directed-MAML is formulated as follows. Given a task distribution \( \rho(\mathcal{T}) \), each task \( \mathcal{T}_i \sim \rho(\mathcal{T}) \) is characterized by a unique environment parameter \( \phi_{\mathcal{T}_i} \in \Phi \), where \( \Phi \) denotes the environment parameter space, the collection of environment parameter $\phi$ which captures key properties or parameters of a specific environment.  We assume that tasks are sampled such that each environment parameter \( \phi_{\mathcal{T}_i} \) is drawn uniformly from environment parameter space \( \Phi \), i.e., \( \phi_{\mathcal{T}_i} \sim \mathcal{U}(\Phi) \). We use \( \mathcal{T}_{\text{med}} \) to denote the task characterized by the mean of the parameters sampled from the task distribution \( \rho(\mathcal{T}) \). The environment parameter \( \phi_{\mathcal{T}_{\text{med}}} \) of the medium task \( \mathcal{T}_{\text{med}} \) is defined in Equation~\ref{formula3}. During meta-RL training, tasks are sampled from a uniformly distributed task distribution \( \rho(\mathcal{T}) \).

\begin{equation}
\label{formula3}
\phi_{\mathcal{T}_{med}} = \mathbb{E}_{\mathcal{T}_i \sim \rho(\mathcal{T})}[\phi_{\mathcal{T}_{i}}].
\end{equation}

If $M$ tasks are sampled from task distribution $\rho(\mathcal{T})$, the medium task can be approximated by:

\begin{equation}
    \phi_{\mathcal{T}_{med}} \approx \frac{1}{M} \sum_{i=1}^{M} \phi_{\mathcal{T}_{i}}.
\end{equation}

\begin{figure}[t]
\centering
\includegraphics[width=1.0\columnwidth]{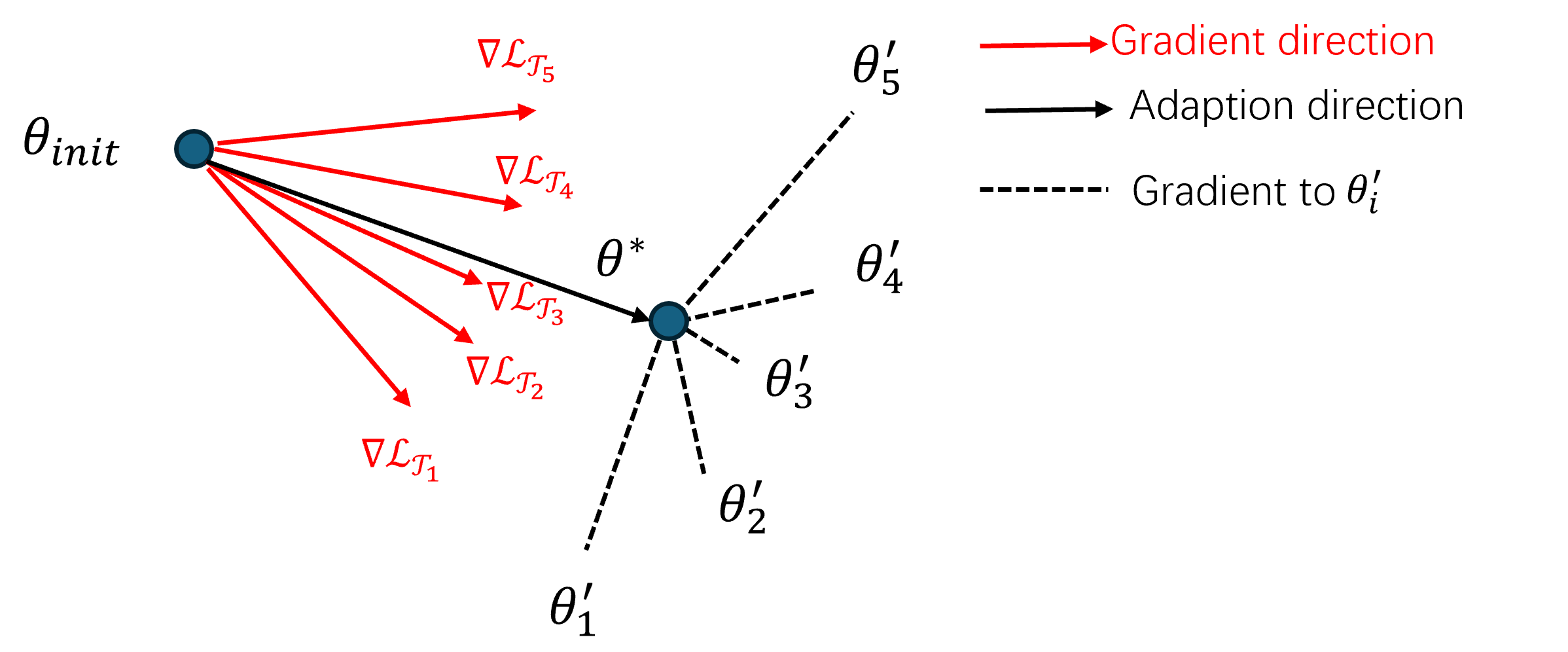}
\caption{One of MAML's outer loop gradient steps on meta-RL ($M=5$): optimal policy parameter $\theta'_i$ for task $\mathcal{T}_{i}$ distributed with $\theta'_3$ (policy parameter of medium task) as the center. The optimal $\theta^{*}$ is approximately consistent with $\theta'_3$ (optimal policy for medium task).}
\label{fig1}
\end{figure}

In MAML, the meta-policy parameters \( \theta \) are updated using the average of task-specific gradients over a batch of sampled tasks (Equation \ref{2b}).
This gradient averaging effect causes the meta-policy to be iteratively pulled toward the region that minimizes the expected loss across tasks. When tasks are sampled such that each environment parameter \( \phi_{\mathcal{T}_i} \) is drawn uniformly from environment parameter space \( \Phi \), the resulting averaged gradient direction tends to align with the optimal policy of the medium task $\mathcal{T}_{\text{med}}$, whose environment parameter lies near the center of the distribution. Intuitively, the meta-policy acts as a compromise among task-specific optima, and thus converges to a point near the geometric center of these optima in parameter space (as shown in Figure \ref{fig1}). This insight provides the motivation for using a representative task—such as the medium task—to approximate or direct the meta-gradient, as done in Directed-MAML. 

\section{Our Method: Directed-MAML}

\begin{algorithm}
\caption{Pseudo-code for Directed-MAML Algorithm.}
\label{alg:directedMAML}
 \textbf{Require:} $\rho(\mathcal{T}):$ Task distribution\\
 \textbf{Require:} $\Phi:$ Environment parameter space\\
 \textbf{Require:} $\alpha,\beta,\delta:$ Step size hyperparameters\\
 \textbf{Require:} $H,E,K:$ Time horizon, Training epoch and Number of trajectories \\
  
\begin{algorithmic}[1]
\State $\phi_{\mathcal{T}_{med}} = \mathbb{E}_{\mathcal{T}_i \sim \rho(\mathcal{T})}[\phi_{\mathcal{T}_{i}}]$
\State Randomly initialize $\theta$
\For{epoch from 1 to $E$}
    \State \parbox[t]{0.85\linewidth}{Sample $K$ trajectories $\mathcal{D}_{med} = \{(s_0^{(l)}, a_0^{(l)}, \dots, s_H^{(l)})\}^{K}_{l=1}$ using $\pi_{\theta}$ in $\mathcal{T}_{\text{med}}$}
    \State Update $\theta \gets \theta + \delta   \nabla_\theta \mathcal{J}_{\mathcal{T}_{med}}(\theta)$  with $\mathcal{D}_{med}$
    \State Sample batch of tasks $\mathcal{T}_i\sim\rho(\mathcal{T})$
    \For{all $\mathcal{T}_i$}
        \State \parbox[t]{0.82\linewidth}{Sample $K$ trajectories $\mathcal{D} = \{(s_0^{(l)}, a_0^{(l)}, \dots, s_H^{(l)})\}^{K}_{l=1}$ using $\pi_{\theta}$ in $\mathcal{T}_{i}$}
        \State Update $\theta_i' \gets \theta + \alpha \nabla_\theta \mathcal{J}_{\mathcal{T}_i}(\theta)$ with $\mathcal{D}$
        \State \parbox[t]{0.83\linewidth}{Resample $K$ trajectories $\mathcal{D'_\text{i}} = \{(s_0^{(l)}, a_0^{(l)}, \dots, s_H^{(l)})\}^{K}_{l=1}$ using $\pi_{\theta'_i}$ in $\mathcal{T}_{i}$}
    \EndFor
    \State \parbox[t]{0.95\linewidth}{Update $\theta \gets \theta + \beta \sum_{\mathcal{T}_i \sim \rho(\mathcal{T})} \nabla_\theta \mathcal{J}_{\mathcal{T}_i}
    (\theta'_i)$ using $\mathcal{D'_\text{i}}$}
\EndFor
\end{algorithmic}
\end{algorithm}

Algorithm~\ref{alg:directedMAML} presents the pseudo-code of Directed-MAML. At the beginning (Line 1), Directed-MAML estimates the environment parameter for medium task $\phi_{\mathcal{T}_{med}}$ by calculating the expectation of environment parameter across task distribution $\rho(\mathcal{T})$. In Line 4, Directed-MAML sample $K$ trajectories from the environment using policy $\pi_\theta$ in the environment defined by $\phi_{\mathcal{T}_{med}}$, and then perform a first-order gradient step using these trajectories to get an early update of $\theta$ in Line 5. This simulates the influence of a second-order term without computing expensive second-order derivative. The step size $\delta$ should be smaller than the outer-loop step size $\beta$ to prevent the meta-policy from overfitting to the medium task $\mathcal{T}_{\text{med}}$. The rest of the algorithm is the same as MAML algorithm on meta-RL.

\begin{figure*}[!t]
    \centering

    \begin{minipage}[b]{0.08\textwidth}
    \end{minipage}%
    \begin{subfigure}[b]{0.32\textwidth}
        \includegraphics[width=\linewidth]{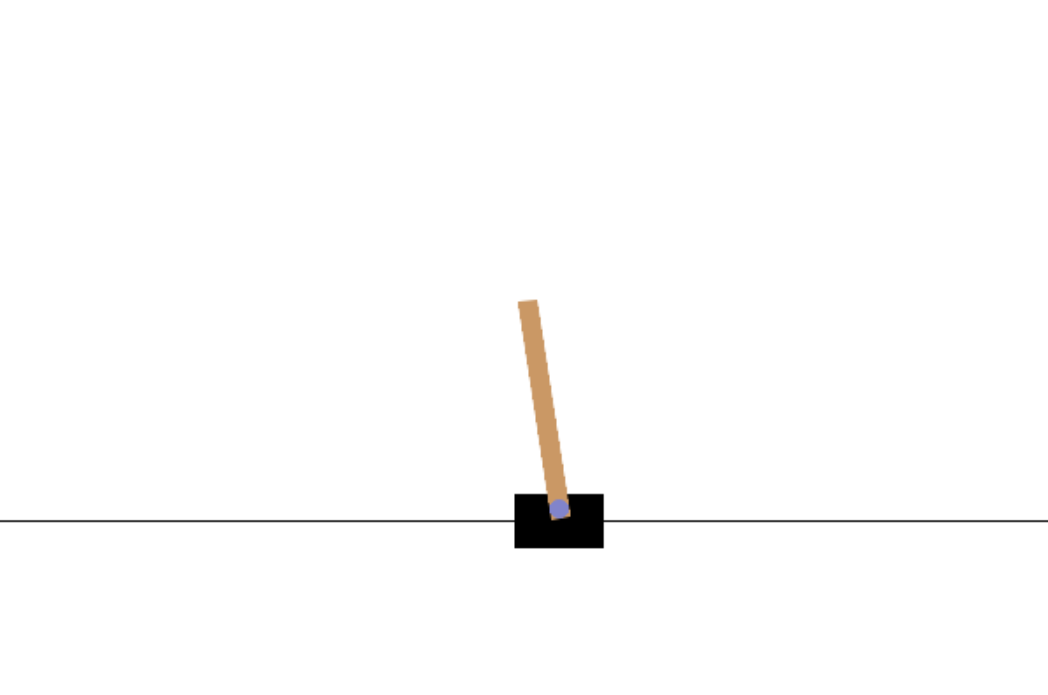}
        \caption{CartPole-v1}
        \label{fig:0}
    \end{subfigure}%
    \begin{subfigure}[b]{0.32\textwidth}
        \includegraphics[width=\linewidth]
        {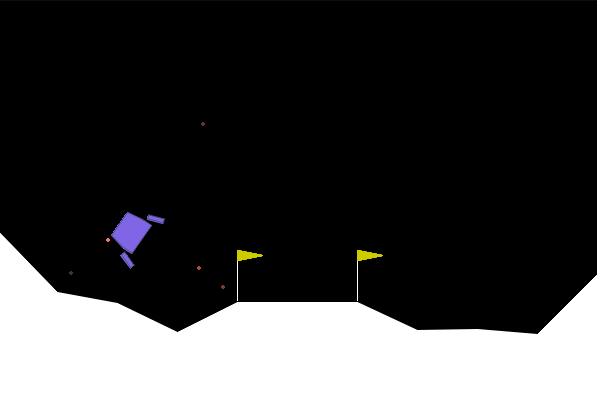}
        \caption{LunarLander-v2}
    \end{subfigure}%
    \begin{subfigure}[b]{0.32\textwidth}
        \includegraphics[width=\linewidth]{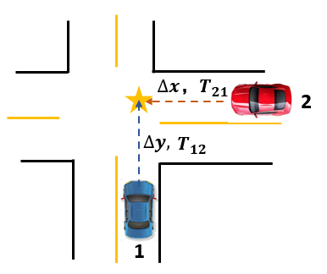}
        \caption{Two-vehicle intersection crossing}
    \end{subfigure}

    \begin{minipage}[b]{0.02\textwidth}
    \end{minipage}%
    \begin{subfigure}[b]{0.32\textwidth}
        \includegraphics[width=\linewidth]{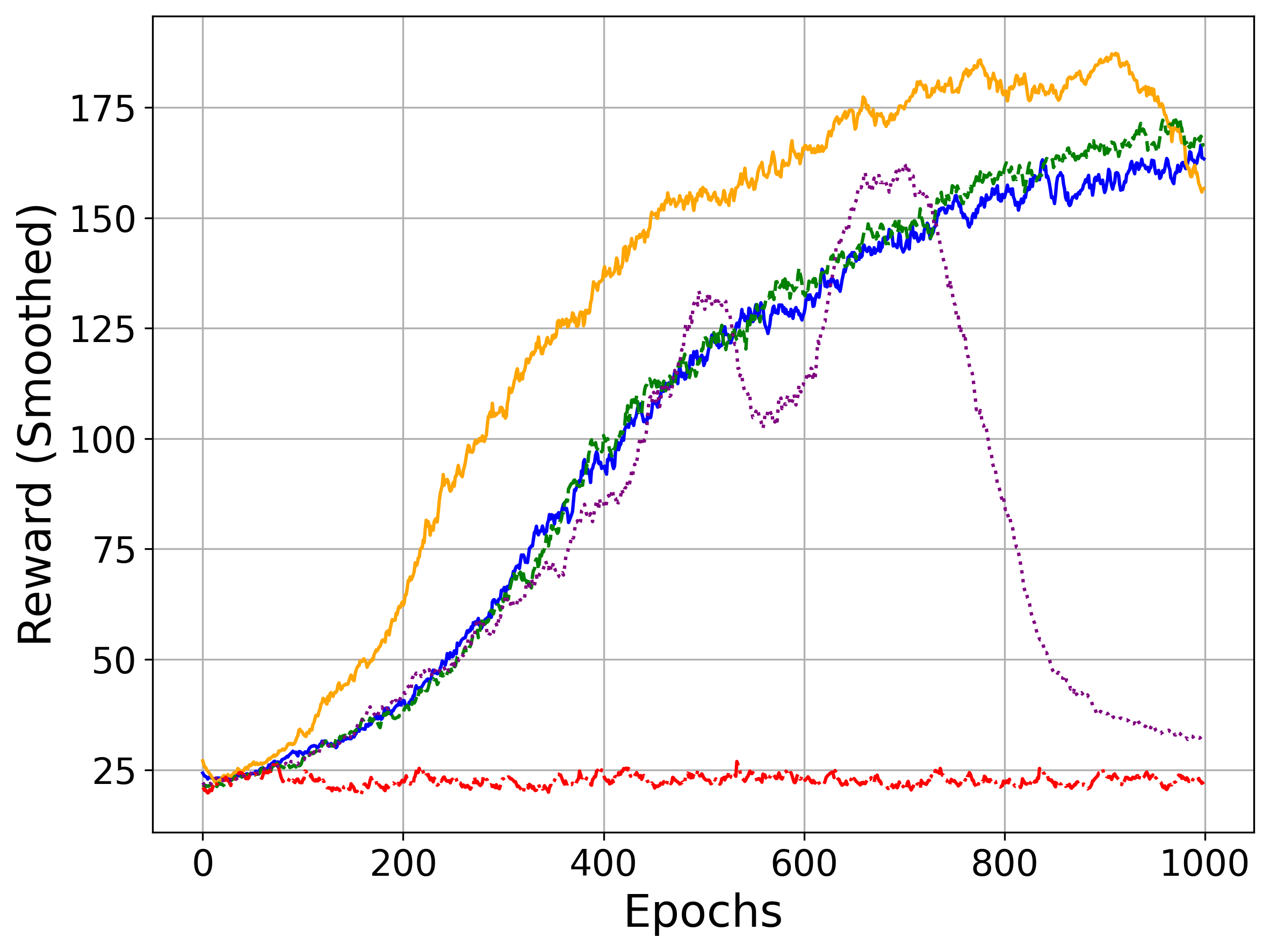}
        \caption{PG on CartPole-v1}
        \label{fig:1}
    \end{subfigure}%
    \begin{subfigure}[b]{0.32\textwidth}
        \includegraphics[width=\linewidth]{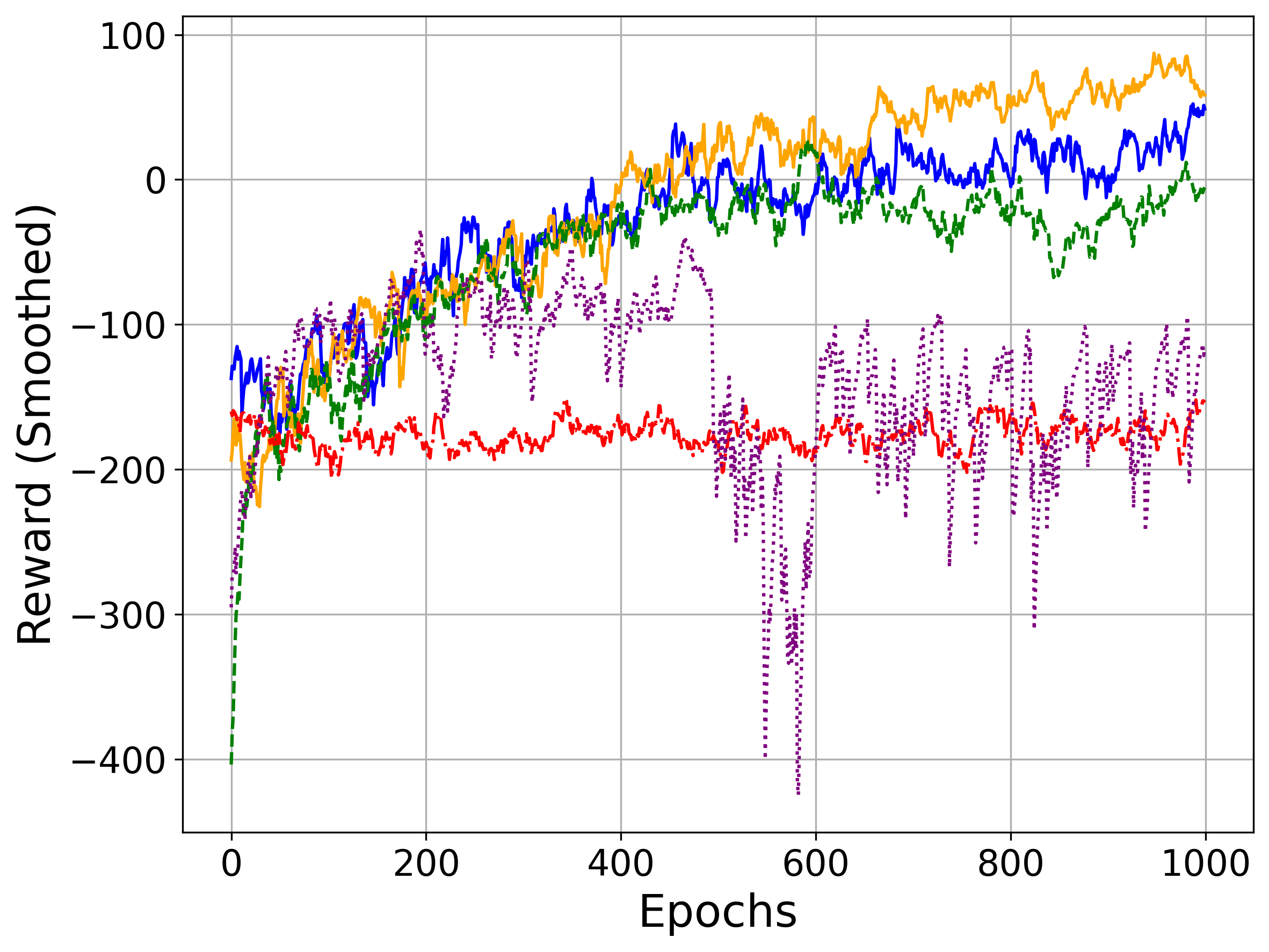}
        \caption{PG on LunarLander-v2}
    \end{subfigure}%
    \begin{minipage}[b]{0.02\textwidth}
    \end{minipage}%
    \begin{subfigure}[b]{0.32\textwidth}
        \includegraphics[width=\linewidth]{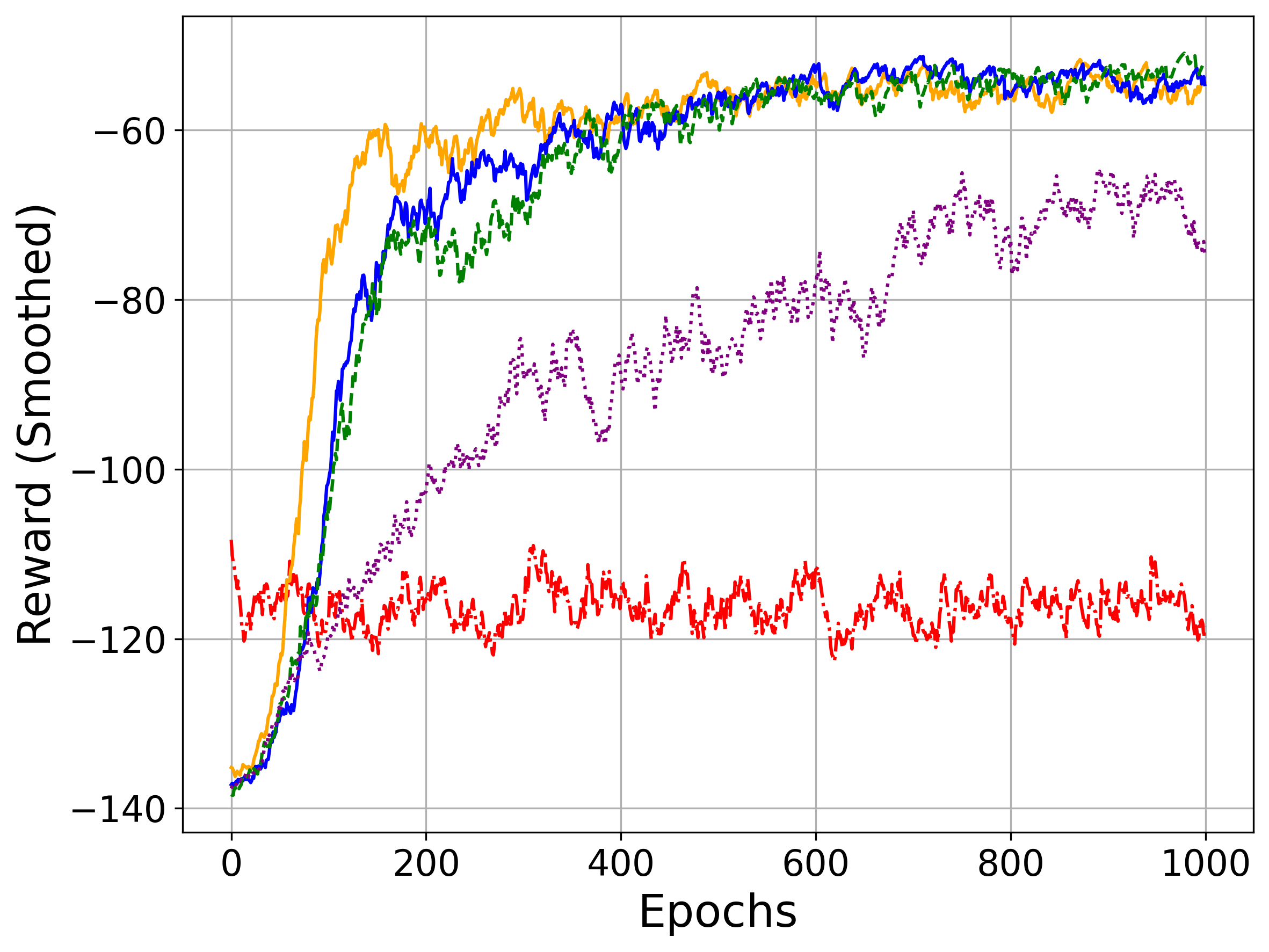}
        \caption{Pg on Two-vehicle intersection crossing}
    \end{subfigure}


    \begin{subfigure}[b]{0.32\textwidth}
        \includegraphics[width=\linewidth]{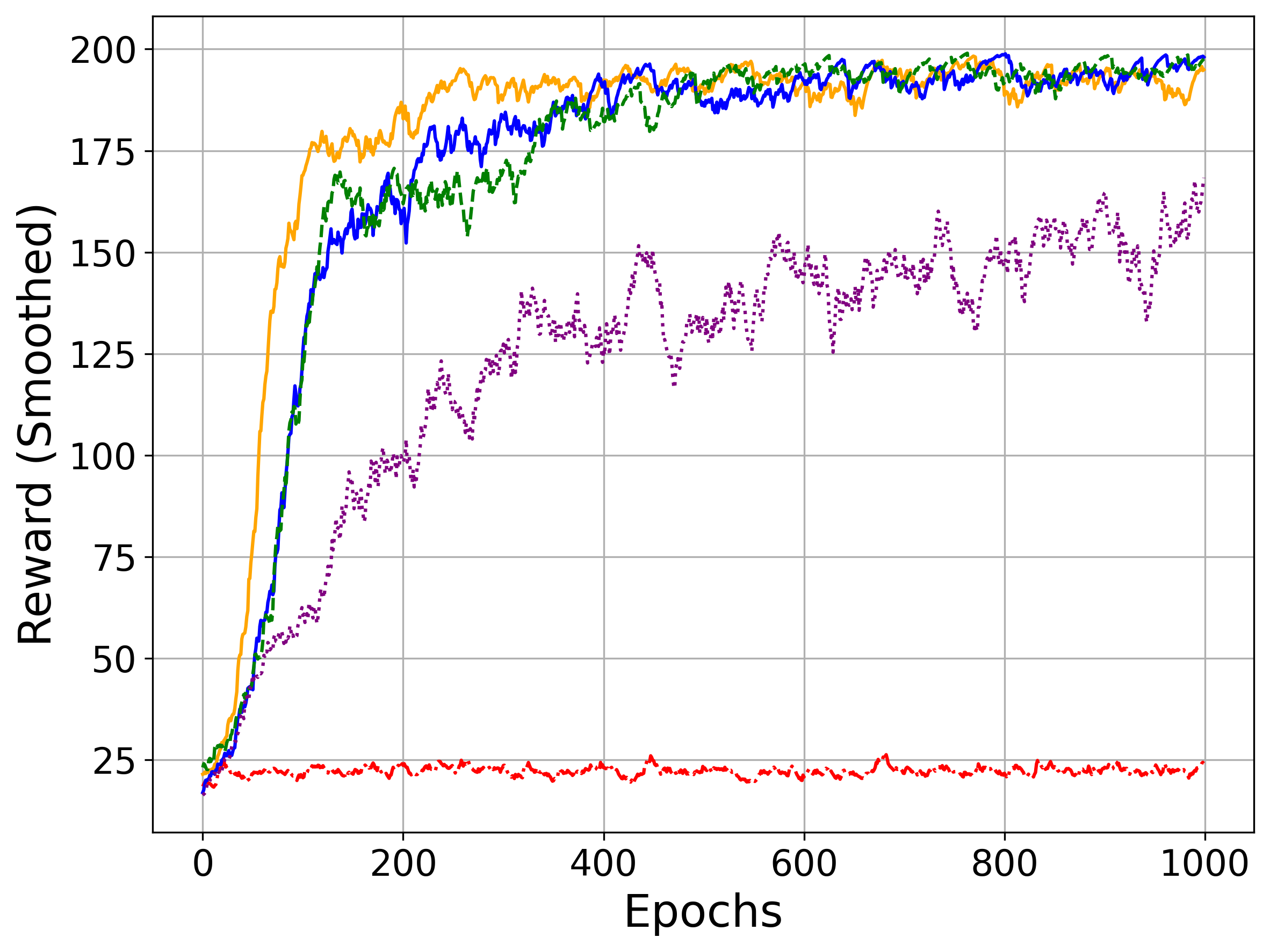}
        \caption{AC on CartPole-v1 }
        \label{fig:2}
    \end{subfigure}%
    \begin{subfigure}[b]{0.32\textwidth}
        \includegraphics[width=\linewidth]{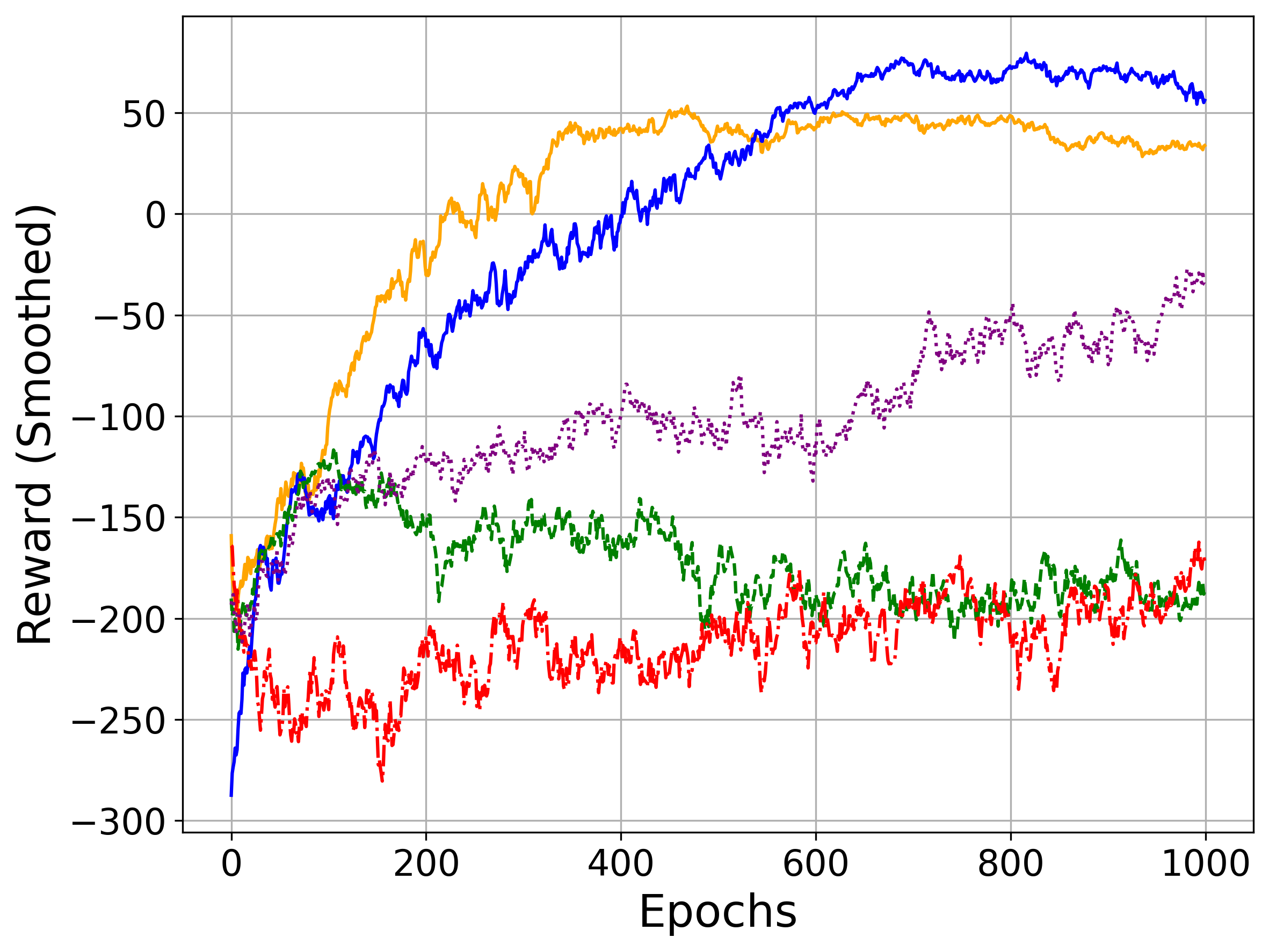}
        \caption{AC on LunarLander-v2}
    \end{subfigure}%
    \begin{subfigure}[b]{0.32\textwidth}
        \includegraphics[width=\linewidth]{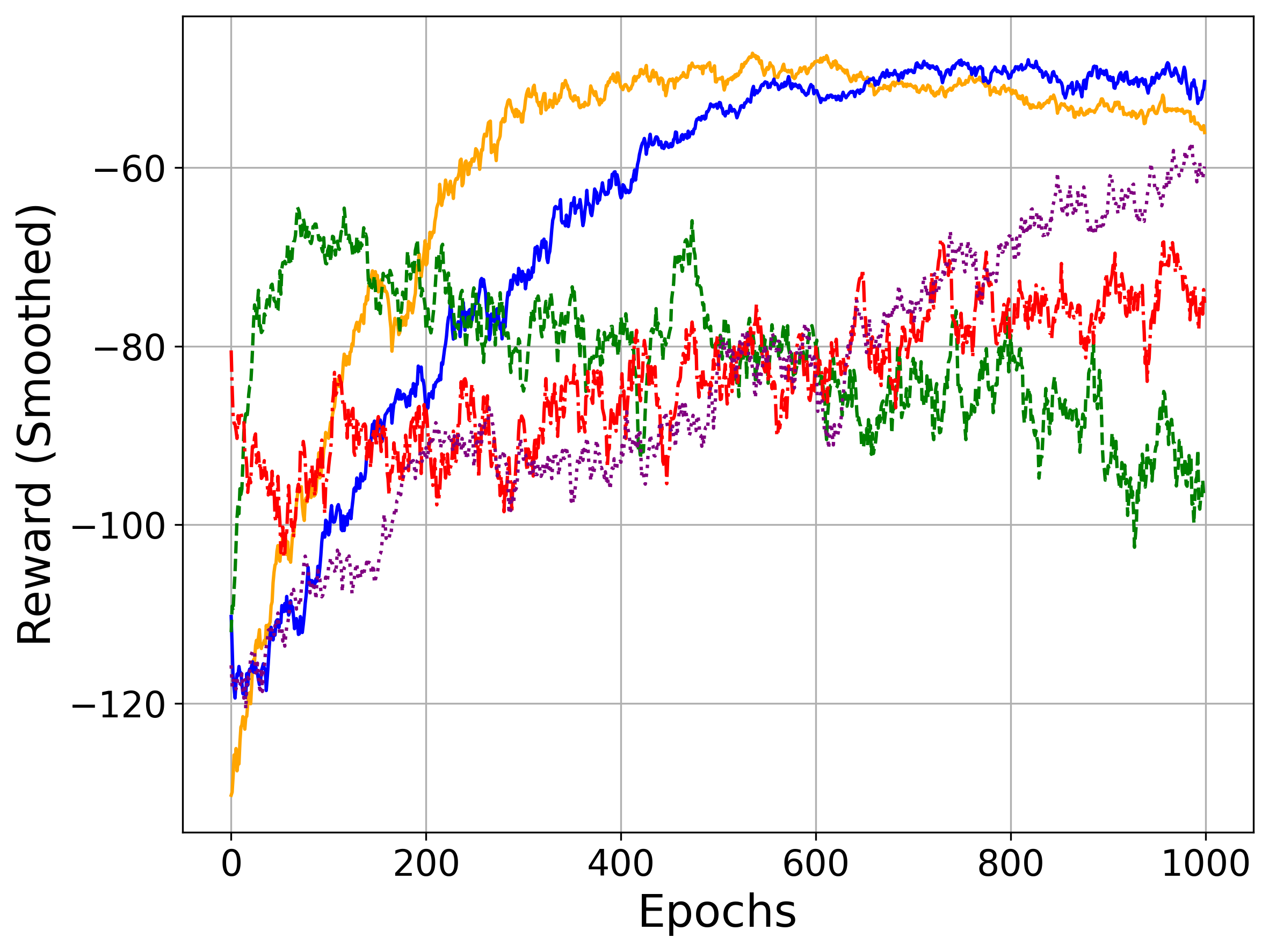}
        \caption{AC on Two-vehicle intersection crossing}
    \end{subfigure}

    \begin{subfigure}[b]{0.9\textwidth}
        \includegraphics[width=\linewidth]{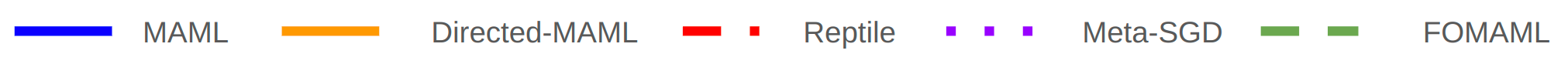}
    \end{subfigure}

    \caption{Performance Comparison of Directed-MAML and Other meta-RL Algorithms on CartPole-v1, LunarLander-v2, and Two-vehicle intersection crossing scenarios. Policy Gradient (PG) and Actor-Critic (AC) methods are used to evaluate performance differences. The yellow curve represents the training performance of Directed-MAML.}
    \label{fig:meta_rl_grid}
\end{figure*}

As discussed in Section 3.2, the medium task $\mathcal{T}_{\text{med}}$ dominates the gradient step in MAML's outer loop. Prior to each gradient step of MAML (both outer loop and inner loop), we compute the first-order gradient of the medium task $\mathcal{T}_{\text{med}}$ and perform one round of gradient step. Task-directed approximation offers three main advantages.

\textbf{Computational efficiency.} 
As an approximation of MAML's second-order derivatives, the task-directed approximation computes only the first-order gradient of the medium task, thereby reducing the computational cost associated with meta-gradient.

\textbf{Global convergence encouragement.}
Standard MAML poses an optimization problem prone to local optima, making convergence to the global optimum difficult. Task-directed approximation guides gradient updates toward the medium task, whose optimal policy is often close to the global meta-policy optimum. This helps MAML escape local optima and improve convergence. However, since the medium task optimum does not always match the true meta-optimum, it may cause instability after convergence.

\textbf{MAML-model agnostic.}
Task-directed approximation is compatible with any MAML-based meta-RL algorithm. It can be incorporated prior to the standard gradient steps of MAML-based methods to guide the optimization. The experiments presented in Section~5 demonstrate the effectiveness of task-directed approximation in enhancing performance.

\section{Experiments}

In this part, we conduct experiments to evaluate the improvements of Directed-MAML in terms of computational efficiency and convergence behavior. Specifically, we consider three tasks—CartPole-v1, LunarLander-v2, and Two-Vehicle Intersection Crossing—and compare the computational efficiency and convergence speed of various meta-learning models. All experiments are conducted on a Dell desktop workstation equipped with an NVIDIA RTX 3090 GPU (24 GB GDDR6X memory). All experiments were conducted on a Dell desktop with an Intel Core i9-12900K CPU, 64 GB RAM and an NVIDIA RTX 3090 GPU (24 GB), running Ubuntu 22.04 LTS with CUDA 12.1 and PyTorch 2.1.

\subsection{Hyperparameter}

The hyperparameter setups are to adjust the learning environment. We apply the same set of hyperparameters for all models in the experiments. The detailed  hyperparameter setups can be found in Table \ref{table1}.

\begin{table}[ht]
\centering
\large 
\caption{Simulation Parameters}
\begin{tabular}{ll}
\toprule
\textbf{Parameters} & \textbf{Value} \\
\midrule
Training epoch $E$ & Vary by task \\
Step size $\delta$ & 0.005 \\
Step size $\alpha$ & 0.001 \\
Step size $\beta$ & 0.001 \\
Discount factor $\gamma$ & 0.99 \\
Sampled task number $M$ & 5 \\
Sampled trajectories number $K$ & 10 \\
Time horizon $H$ & Vary by task \\
\bottomrule
\end{tabular}
\label{table1}
\end{table}

\subsection{Experiment Scenarios}

To evaluate the performance of Directed-MAML compared with other MAML-based algorithms, we tested their performance on two OpenAI Gym scenarios \cite{openaigym}, CartPole-v1 and LunarLander-v2, and a two-vehicle intersection crossing scenario. It should be noticed that the transition model $\mathcal{P}_i$ for task $\mathcal{T}_i$ is deterministic in our setup for Two-vehicle intersection crossing scenario: For any state-action pair $(s, a)$, the resulting next state $s'$ is uniquely determined, i.e., $\mathcal{P}_i(s' \mid s, a) = 1$.

\textbf{CartPole-v1:}
The CartPole environment is a standard reinforcement learning benchmark in which the goal is to balance a pole on a moving cart by applying discrete forces. The state space is continuous and defined as $s = (x, \dot{x}, \theta, \dot{\theta})$, representing the cart position and velocity, and the pole angle and angular velocity. The action space is discrete, with $a \in \{0, 1\}$ corresponding to pushing the cart left or right.  The agent receives a reward of $+1$ for each time step the pole remains upright, and an episode (up to 200 steps) terminates if the pole falls or the cart moves out of bounds. For meta-RL on the CartPole-v1 task, each task $\mathcal{T}_i \sim \rho(\mathcal{T})$ represents an individual MDP with different gravity. Specifically, the gravity of task $\mathcal{T}_i$ is sampled from the environment parameter space $\Phi = [5.0, 15.0]$ m/s$^2$.

\textbf{LunarLander-v2:} 
    The LunarLander-v2 environment is a physics-based reinforcement learning benchmark in which the goal is to control a lunar module to land safely on a designated landing pad. The state space is continuous and defined as $s = (x, y, \dot{x}, \dot{y}, \theta, \dot{\theta}, c_l, c_r)$, where $(x, y)$ denote the lander's position, $(\dot{x}, \dot{y})$ its velocity, $\theta$ and $\dot{\theta}$ its angle and angular velocity, and $c_l$, $c_r$ are binary indicators for whether the left or right leg is in contact with the ground. The action space is discrete with $a \in \{0, 1, 2, 3\}$, corresponding to: do nothing, fire left engine, fire main engine, or fire right engine. The agent receives rewards based on its distance to the landing pad, speed, fuel usage, and leg contact, with a successful landing yielding a total reward of approximately $+200$, and crashing resulting in a large negative reward. For meta-RL, each task $\mathcal{T}_i \sim \rho(\mathcal{T})$ corresponds to an MDP with varied gravity. Specifically, the gravity of task $\mathcal{T}_i$ is sampled from the environment parameter space $\Phi = [5.0, 15.0]$ m/s$^2$.

\textbf{Two-vehicle intersection crossing:}
We consider a vehicle control environment where the objective is to learn a policy for Vehicle~1 to maintain an optimal speed and safe distance from Vehicle~2, which moves with a fixed velocity $u$\,m/s. The state is defined as $s = (\Delta x, \Delta y)$, representing the relative position between the two vehicles. The action space is continuous, and $a \in [0, 15]$\,m/s, representing the velocity command for Vehicle~1. The policy $\pi : \mathcal{S} \to \mathcal{A}$ is parameterized over a  space which is continuous. For meta-RL, each task $\mathcal{T}_i \sim \rho(\mathcal{T})$ corresponds to an MDP with varied Vehicle 2 speeds. Specifically,  the Vehicle~2 speed of $\mathcal{T}_i$ is sampled from the environment parameter space $\Phi = [5.0, 15.0]$m/s.

\subsection{Experiment Results}

\begin{table*}[t]
\centering
\caption{Computational comparison (mean $\pm$ std, 5 seeds) between Directed-MAML and MAML-based algorithms on LunarLander-v2.}
\begin{tabular}{lccc}
\toprule
Method & Runtime for one epoch (s) & Runtime for 1000 epochs (h) & Runtime to convergence (h) \\
\midrule
MAML & $2.34 \pm 0.12$ & $0.66 \pm 0.09$ & $0.39 \pm 0.06$ \\
FOMAML & $1.56 \pm 0.08$ & $0.43 \pm 0.05$ & $0.43 \pm 0.05$ \\
Meta-SGD & $\textbf{1.41} \pm \textbf{0.03}$ & $\textbf{0.38} \pm \textbf{0.04}$ & $0.38 \pm 0.08$ \\
Directed-MAML & $2.52 \pm 0.09 $ & $0.71 \pm 0.05$ & $\textbf{0.22} \pm \textbf{0.03}$ \\
\midrule
Speedup (MAML / Directed-MAML) & 0.93$\times$ & 0.93 $\times$ & 1.77$\times$  \\
\bottomrule
\end{tabular}
\label{tab:computationcost}
\end{table*}

\subsubsection{Directed-MAML v.s. other meta-RL models}

We compare the required convergence epochs of Directed MAML with other gradient-based meta-RL methods on three scenarios: CartPole-v1, LunarLander-v2, and Two-vehicle intersection crossing scenario. The baseline methods include MAML~\cite{MAML}, Reptile~\cite{reptile}, Meta-SGD~\cite{metasgd}, and FOMAML~\cite{FOMAML}, all of which are gradient-based meta-learning algorithms for fair comparison. We evaluate each model using two reinforcement learning approaches: policy gradient and actor-critic algorithms. To ensure consistency and reproducibility, all models share the same hyperparameter configurations, which are initialized with the same policy parameters $\theta$ and are trained using the same random seed (seed = 1).

Figure~\ref{fig:meta_rl_grid} illustrates the training curves of all experimental models. For better visualization, all curves are smoothed using an exponential moving average (EMA) with a smoothing factor of 0.9. As shown in the figure, Directed MAML outperforms other gradient-based meta-RL algorithms in terms of required convergence epochs for all experiment setups. Taking Figure~\ref{fig:2} as an example, Among the evaluated meta-RL algorithms, Directed-MAML achieves the fastest and most stable convergence, reaching a smoothed reward above 175 by around epoch 150 and approaching the optimal reward of 200 by epoch 300. MAML also demonstrates strong performance, surpassing 175 reward around epoch 250 and steadily converging just below Directed-MAML by approximately epoch 400. Both methods exhibit stable learning dynamics, with Directed-MAML consistently maintaining a performance advantage throughout training. Considering the reduced computational cost of task-directed approximation compared with second-order gradient, Directed-MAML is significantly more efficient in terms of computation.

\subsubsection{Comparison of computational efficiency.}

Table~\ref{tab:computationcost} compares the computational efficiency of MAML, FOMAML, Meta-SGD, and Directed-MAML on LunarLander-v2. Although Directed-MAML has a slightly higher per-epoch runtime than MAML (2.52 s vs. 2.34 s), it reaches convergence much faster, requiring only 0.22 h compared to 0.39 h for MAML. This corresponds to a 1.77× speedup in convergence time, demonstrating that the task-directed approximation effectively reduces the number of training epochs needed despite the added per-epoch cost. While FOMAML and Meta-SGD remain the most efficient in terms of per-epoch runtime, Directed-MAML surpasses them in terms of runtime to convergence, highlighting its practical computational advantage.

\subsubsection{Task-directed approximation on Meta-SGD and FOMAML.}

\begin{figure}[ht]
\centering
\includegraphics[width=0.80\columnwidth]{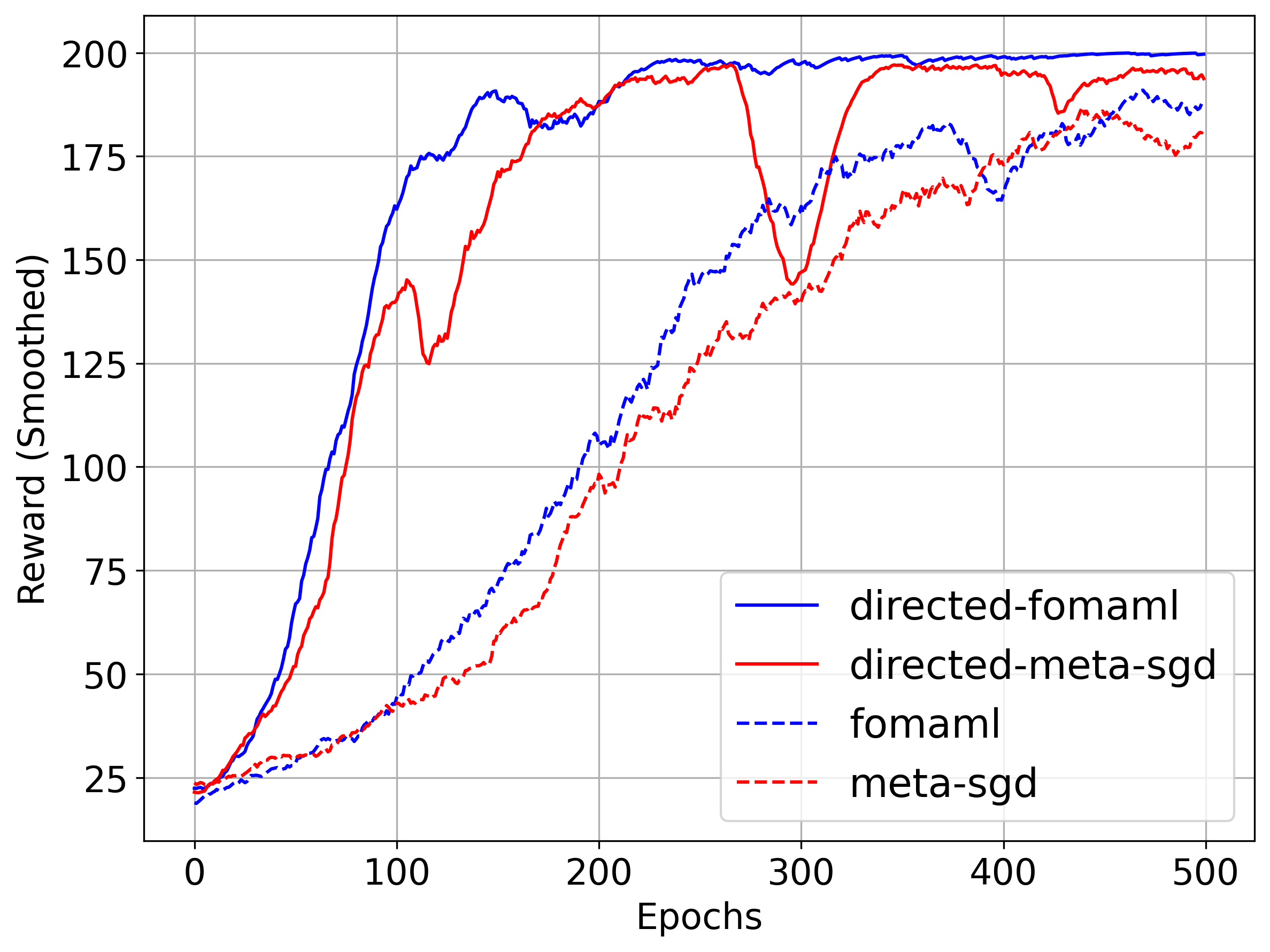}
\caption{Training Convergence Comparison of Directed-FOMAML and Directed-Meta-SGD on CartPole-v1 scenario (Step Size $\beta =0.02$) }
\label{fig:directed_variants}
\end{figure}

Since the task-directed approximation is model-agnostic and compatible with any MAML-based meta-learning algorithm, we further explored its applicability beyond MAML by integrating it into Meta-SGD and FOMAML. In these variants, the task-directed approximation is applied prior to the gradient update step during meta-training. The resulting modified algorithms are referred to as Directed-Meta-SGD and Directed-FOMAML, respectively. As shown in Figure~\ref{fig:directed_variants}, Both directed-fomaml and directed-meta-sgd demonstrate faster convergence and higher final performance compared to their non-directed baselines, validating the effectiveness of the task-directed approximation in improving sample efficiency and stability in meta-reinforcement learning. However, we should also note that the noticeable turbulence observed after the training curve converges in Directed-FOMAML and Directed-Meta-SGD originates from the task-directed approximation.

\section{Conclusions and future work}

In this paper, we propose Directed-MAML, a novel gradient-based meta-RL algorithm that incorporates a task-directed approximation strategy into MAML to improve both computational efficiency and convergence speed. Directed-MAML demonstrates superior performance compared to existing gradient-based meta-learning algorithms, requiring fewer epochs to converge while maintaining high-quality policy adaptation across meta-RL tasks. Experimental results show that Directed-MAML achieves faster overall runtime to convergence, outperforming other meta-RL methods in practical efficiency. Furthermore, the task-directed approximation represents a generalizable strategy that can be incorporated into other gradient-based meta-RL algorithms, such as Meta-SGD and FOMAML, leading to similar improvements in efficiency and reduced training epochs.

Although Directed-MAML currently relies on a uniform task sampling strategy , extending it to handle more diverse or unstructured task distributions provides a promising avenue for future research. Additionally, minor fluctuations observed after convergence suggest sensitivity to medium-difficulty tasks; addressing these post-convergence dynamics may further enhance the stability and robustness of Directed-MAML. Overall, our work demonstrates that task-directed approximation is an effective and flexible approach for improving both the computational efficiency and convergence properties of gradient-based meta-RL algorithms.


\bibliographystyle{IEEEtran}
\bibliography{ref}

\begin{thebibliography}{10}
\providecommand{\url}[1]{#1}
\csname url@samestyle\endcsname
\providecommand{\newblock}{\relax}
\providecommand{\bibinfo}[2]{#2}
\providecommand{\BIBentrySTDinterwordspacing}{\spaceskip=0pt\relax}
\providecommand{\BIBentryALTinterwordstretchfactor}{4}
\providecommand{\BIBentryALTinterwordspacing}{\spaceskip=\fontdimen2\font plus
\BIBentryALTinterwordstretchfactor\fontdimen3\font minus \fontdimen4\font\relax}
\providecommand{\BIBforeignlanguage}[2]{{%
\expandafter\ifx\csname l@#1\endcsname\relax
\typeout{** WARNING: IEEEtran.bst: No hyphenation pattern has been}%
\typeout{** loaded for the language `#1'. Using the pattern for}%
\typeout{** the default language instead.}%
\else
\language=\csname l@#1\endcsname
\fi
#2}}
\providecommand{\BIBdecl}{\relax}
\BIBdecl

\bibitem{zhang2017understanding}
C.~Zhang, S.~Bengio, M.~Hardt, B.~Recht, and O.~Vinyals, ``Understanding deep learning requires rethinking generalization,'' in \emph{International Conference on Learning Representations (ICLR)}, 2017.

\bibitem{shorten2019survey}
C.~Shorten and T.~M. Khoshgoftaar, ``A survey on image data augmentation for deep learning,'' \emph{Journal of Big Data}, vol.~6, no.~1, pp. 1--48, 2019.

\bibitem{DBLP:journals/corr/abs-1909-13579}
E.~Bennequin, ``Meta-learning algorithms for few-shot computer vision,'' \emph{CoRR}, vol. abs/1909.13579, 2019.

\bibitem{DBLP:journals/corr/abs-2007-09604}
W.~Yin, ``Meta-learning for few-shot natural language processing: A survey,'' \emph{CoRR}, vol. abs/2007.09604, 2020.

\bibitem{DBLP:journals/corr/abs-2009-08445}
T.~Bansal, R.~Jha, T.~Munkhdalai, and A.~McCallum, ``Self-supervised meta-learning for few-shot natural language classification tasks,'' \emph{CoRR}, vol. abs/2009.08445, 2020.

\bibitem{Gui_2018_ECCV}
L.-Y. Gui, Y.-X. Wang, D.~Ramanan, and J.~M.~F. Moura, ``Few-shot human motion prediction via meta-learning,'' in \emph{Proceedings of the European Conference on Computer Vision (ECCV)}, September 2018.

\bibitem{Elsken_2020_CVPR}
T.~Elsken, B.~Staffler, J.~H. Metzen, and F.~Hutter, ``Meta-learning of neural architectures for few-shot learning,'' in \emph{Proceedings of the IEEE/CVF Conference on Computer Vision and Pattern Recognition (CVPR)}, June 2020.

\bibitem{lee-etal-2021-meta}
H.-y. Lee, N.~T. Vu, and S.-W. Li, ``Meta learning and its applications to natural language processing,'' in \emph{Proceedings of the 59th Annual Meeting of the Association for Computational Linguistics and the 11th International Joint Conference on Natural Language Processing: Tutorial Abstracts}, D.~Chiang and M.~Zhang, Eds.\hskip 1em plus 0.5em minus 0.4em\relax Online: Association for Computational Linguistics, Aug. 2021, pp. 15--20.

\bibitem{Wang_2019_ICCV}
Y.-X. Wang, D.~Ramanan, and M.~Hebert, ``Meta-learning to detect rare objects,'' in \emph{Proceedings of the IEEE/CVF International Conference on Computer Vision (ICCV)}, October 2019.

\bibitem{Wang_2022_CVPR}
H.~Wang, Y.~Wang, R.~Sun, and B.~Li, ``Global convergence of maml and theory-inspired neural architecture search for few-shot learning,'' in \emph{Proceedings of the IEEE/CVF Conference on Computer Vision and Pattern Recognition (CVPR)}, June 2022, pp. 9797--9808.

\bibitem{MAML}
C.~Finn, P.~Abbeel, and S.~Levine, ``Model-agnostic meta-learning for fast adaptation of deep networks,'' in \emph{Proceedings of the 34th International Conference on Machine Learning}, ser. Proceedings of Machine Learning Research, D.~Precup and Y.~W. Teh, Eds., vol.~70.\hskip 1em plus 0.5em minus 0.4em\relax PMLR, 06--11 Aug 2017, pp. 1126--1135.

\bibitem{mamlcv1}
J.~Chen and C.~Deng, ``Maml mot: Multiple object tracking based on meta-learning,'' in \emph{2024 IEEE International Conference on Systems, Man, and Cybernetics (SMC)}, 2024, pp. 4542--4547.

\bibitem{mamlcv2}
S.~Khodadadeh, L.~Boloni, and M.~Shah, ``Unsupervised meta-learning for few-shot image classification,'' in \emph{Advances in Neural Information Processing Systems}, H.~Wallach, H.~Larochelle, A.~Beygelzimer, F.~d\textquotesingle Alch\'{e}-Buc, E.~Fox, and R.~Garnett, Eds., vol.~32.\hskip 1em plus 0.5em minus 0.4em\relax Curran Associates, Inc., 2019.

\bibitem{mamlnlp1}
J.~R. Garcia, F.~Freddi, F.-T. Liao, J.~McGowan, T.~Nieradzik, D.~shan Shiu, Y.~Tian, and A.~Bernacchia, ``Meta-learning with maml on trees,'' 2021.

\bibitem{sharpMAML}
S.~Ravi, A.~Beatson, C.~Zhang, and R.~Ranganath, ``Amortized bayesian meta-learning with sharpness-aware minimization,'' in \emph{International Conference on Learning Representations}, 2019.

\bibitem{sutton2018reinforcement}
R.~S. Sutton and A.~G. Barto, \emph{Reinforcement learning: An introduction}, 2nd~ed.\hskip 1em plus 0.5em minus 0.4em\relax MIT press, 2018.

\bibitem{qlearning}
C.~J. Watkins and P.~Dayan, ``Q-learning,'' \emph{Machine Learning}, vol.~8, no. 3-4, pp. 279--292, 1992.

\bibitem{dqn}
V.~Mnih, K.~Kavukcuoglu, D.~Silver, A.~A. Rusu, J.~Veness, M.~G. Bellemare, A.~Graves, M.~Riedmiller, A.~K. Fidjeland, G.~Ostrovski \emph{et~al.}, ``Human-level control through deep reinforcement learning,'' \emph{Nature}, vol. 518, no. 7540, pp. 529--533, 2015.

\bibitem{ac}
V.~R. Konda and J.~N. Tsitsiklis, ``Actor-critic algorithms,'' Massachusetts Institute of Technology, Tech. Rep., 2000.

\bibitem{rl2}
Y.~Duan, J.~Schulman, X.~Chen, P.~Bartlett, I.~Sutskever, and P.~Abbeel, ``{RL}$^2$: Fast reinforcement learning via slow reinforcement learning,'' \emph{arXiv preprint arXiv:1611.02779}, 2016.

\bibitem{snail}
N.~Mishra, M.~Rohaninejad, X.~Chen, and P.~Abbeel, ``A simple neural attentive meta-learner,'' in \emph{International Conference on Learning Representations}, 2018.

\bibitem{pearl}
K.~Rakelly, A.~Zhou, D.~Quillen, C.~Finn, and S.~Levine, ``Efficient off-policy meta-reinforcement learning via probabilistic context variables,'' in \emph{Proceedings of the 36th International Conference on Machine Learning}.\hskip 1em plus 0.5em minus 0.4em\relax PMLR, 2019, pp. 5331--5340.

\bibitem{mamlplus}
A.~Antoniou, H.~Edwards, and A.~Storkey, ``How to train your maml,'' \emph{arXiv preprint arXiv:1810.09502}, 2019.

\bibitem{FOMAML}
A.~Nichol, J.~Achiam, and J.~Schulman, ``On first-order meta-learning algorithms,'' \emph{arXiv preprint arXiv:1803.02999}, 2018.

\bibitem{reptile}
A.~Nichol and J.~Schulman, ``Reptile: A scalable meta-learning algorithm,'' \emph{arXiv preprint arXiv:1803.02999}, 2018, presented at the OpenAI blog in March 2018.

\bibitem{iMAML}
A.~Rajeswaran, C.~Finn, S.~Kakade, and S.~Levine, ``Meta-learning with implicit gradients,'' in \emph{Advances in Neural Information Processing Systems}, 2019.

\bibitem{openaigym}
G.~Brockman, V.~Cheung, L.~Pettersson, J.~Schneider, J.~Schulman, J.~Tang, and W.~Zaremba, ``Openai gym,'' 2016.

\bibitem{metasgd}
Z.~Li, F.~Zhou, F.~Chen, and H.~Li, ``Meta-sgd: Learning to learn quickly for few-shot learning,'' \emph{arXiv preprint arXiv:1707.09835}, 2017.

\end{thebibliography}

\end{document}